# WS-SfMLearner: Self-supervised Monocular Depth and Ego-motion Estimation on Surgical Videos with Unknown Camera Parameters

Ange Lou[1], Jack Noble[1]

Department of Electrical Engineering[1], Vanderbilt University, Nashville TN, USA

{ange.lou, jack.noble}@vanderbilt.edu

## Abstract

Depth estimation in surgical video plays a crucial role in many image-guided surgery procedures. However, it is difficult and time consuming to create depth map ground truth datasets in surgical videos due in part to inconsistent brightness and noise in the surgical scene. Therefore, building an accurate and robust self-supervised depth and camera ego-motion estimation system is gaining more attention from the computer vision community. Although several self-supervision methods alleviate the need for ground truth depth maps and poses, they still need known camera intrinsic parameters, which are often missing or not recorded. Moreover, the camera intrinsic prediction methods in existing works depend heavily on the quality of datasets. In this work, we aimed to build a self-supervised depth and ego-motion estimation system which can predict not only accurate depth maps and camera pose, but also camera intrinsic parameters. We proposed a cost-volume-based supervision manner to give the system auxiliary supervision for camera parameters prediction. The experimental results showed that the proposed method improved the accuracy of estimated camera parameters, ego-motion, and depth estimation.

**Keywords:** Self-supervised learning, Unknown camera, Depth estimation, Ego-motion estimation

## 1. INTRODUCTION

Depth estimation from surgical images plays an important role in the field of image-guided surgery, such as 3D reconstruction [32], navigation, and augmented reality (AR). Depth estimation from surgical video frames is a challenging task due to several factors such as non-Lambertian reflection properties of tissues, motion blur, and lack of photometric consistency across frames [1]. However, recent advances in computer vision and deep learning techniques have shown promising results in addressing these challenges. Two popular approaches for depth estimation from a sequence of single camera surgical images are structure-from-motion (SfM) [2] and shape-from-shading (SfS) [3], which are two approaches that use classical stereo reconstruction methods between frames acquired at two time points from two different camera positions relative to the scene. Another approach is to use monocular depth estimation, which relies on learning-based methods to infer a depth map of the scene from a single image.

Recently, deep neural networks have shown effectiveness in monocular video frame depth estimation. Fully supervised approaches, e.g., Xu et al [4], Cao et al [5], and Fu et al [6], have achieved outstanding results. Unfortunately, it is difficult to collect large-scale and accurate surgical video depth map datasets due to the difficulty in creating a ground truth. Thus, few such public datasets are available. As a result, many promising self-supervised monocular depth and camera ego-motion estimation networks are proposed to solve this dilemma, such as SfMLearner [7], Monodepth2 [8], SC-SfMLearner [9] and Endo-SfM [10]. These self-supervised methods use disparity information between adjacent frames to supervise the neural networks and can produce relative depth maps. Relative depth maps do not encode exact depth but rather the relative distance to the camera between different objects in the scene and have been shown to improve 3D scene understanding [31]. The common element of these SfM-inspired neural network systems is to predict the depth and ego-motion simultaneously and then to warp source frames to target frames using the coordinate transformation represented in equation 1:

$$h(\boldsymbol{p}^{t\rightarrow s}) = [\boldsymbol{K}|\boldsymbol{0}]M^{t\rightarrow s}\begin{bmatrix}\boldsymbol{D}^t\boldsymbol{K}^{-1}h(\boldsymbol{p}^t)\\1\end{bmatrix}, \qquad (1)$$

where $h(\boldsymbol{p}^t)$ and $h(\boldsymbol{p}^{t\rightarrow s})$ are the homogeneous pixel coordinates in target frame $t$ and those from target frame $t$ mapped to source frame $s$; and $\boldsymbol{K}$, $\boldsymbol{D}^t$, and $\boldsymbol{M}^{t\rightarrow s}$ are transformations representing the camera intrinsic parameters, the depth map of target frame, and the camera ego-motion from $t$ to $s$, respectively. The architecture of this type of neural network system is shown in Fig. 1.

Various types of SfM learner systems have successfully solved the major challenges of surgical video analysis, such as motion blur, specular reflection, and regions with homogeneous content. However, to achieve promising performance by unsupervised methods, a large dataset is crucial. There are many surgical video datasets [11][12] that do not provide the camera intrinsic parameters or camera motion data, which prevent using those data to train existing unsupervised depth estimation networks. Gordon et al [13] overcame this challenge when using unknown source data from natural scenes by first using a convolutional neural network (CNN) to predict the camera intrinsic parameters from unlabeled data. Although Gordon's method achieves promising results on depth map and camera intrinsic parameters estimation, the accuracy of predicted camera intrinsic parameters was shown to be highly dependent on the dataset itself. For example, a dataset with more camera rotations enables more accurate camera intrinsic prediction [13]. Therefore, we hypothesize that a camera intrinsic self-supervision strategy can improve the depth and ego-motion estimation performance. However, it is more challenging to predict accurate depth maps and camera intrinsic parameters in the unknown surgical scene than in natural images due to the limitation of labeled data and the challenges involved with surgical video analysis discussed above.

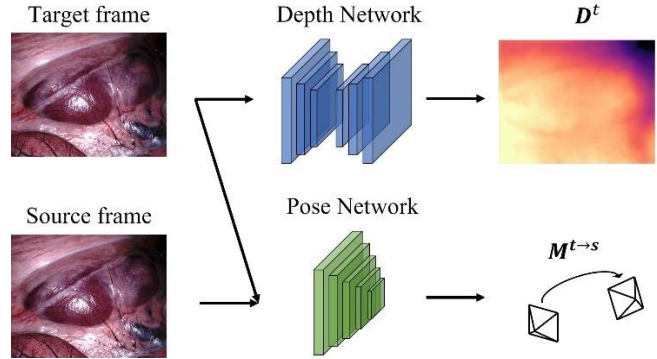

Fig.1. The SfM-inspired system consists of a Depth Network and Pose Network that used to predict depth map for target frame and motion (pose) from target to source, respectively.

Our contributions are as follows: To the best of our knowledge, our proposed method, which we call "WS-SfMLearner" for Wild, Surgical SfM-Learner, is the first that estimates the relative depth of the surgical scene and camera ego-motion using monocular camera video frames with unknown camera positions and intrinsic parameters. Further, we design a cost-volume-based camera intrinsic parameter self-supervision method to train a CNN-based camera intrinsic parameter prediction module. This leads to improved relative depth and ego-motion estimate with WS-SfMLearner. Because WS-SfMLearner does not rely on known camera positions, intrinsic parameters, or stereo images, it is widely applicable to many surgical video datasets.

## 2. RELATED WORK

In this section, we review relevant self-supervised deep learning-based methods in depth estimation. SfMLearner [7] developed the self-supervised framework to solve the unsupervised depth prediction problem as a warping-based view syn-thesis task. After that, many works improve the performance of SfMLearner by exploring geometric constraints [14][15], designing representation learning com-ponents [16][17], and applying attention mechanisms [18]. Although these meth-ods work well in autonomous driving area, they are not generally applicable to surgical environments.

Recently, Turan et al [19] first proposed a self-supervised depth and ego-motion estimation system in endoscope scenes. After that, many works exploring methods to enhance the photometric robustness such as Endo-SfMLearner [10] used an affine brightness transformer and AF-SfMLearner [20] introduced appearance flow. However, seldom studies solve the longstanding challenge – unknown source videos depth and ego-motion estimation. Gordon et al [13] first solved this problem by introducing a camera intrinsic CNN, which predicts camera parameters during training. However, later works [21][22] do not investigate self-supervision methods to improve camera intrinsic estimation.

## 3. METHOD

In this section, we will first introduce the baseline self-supervised system we used – AF-SfMLearner. Then we show the details of the camera intrinsic prediction CNN we propose. Finally, we introduce the cost-volume-based method to supervise the camera CNN training.

### 3.1. Baseline: AF-SfMLearner

AF-SfMLearner as shown in Fig. 2 contains a depth network (DepthNet) and pose network (PoseNet) to predict the depth map of the target frame ($I^t(\boldsymbol{p})$) and ego-motion from the target to the source frame ($I^s(\boldsymbol{p})$). In addition, AF-SfMLearner also uses OFNet to predict the optical flow between target and source frame. The optical flow is used to generate the source-to-target frame and the visibility mask. And then the original target frame and source-to-target are sent into AFNet to do brightness calibration to generate a refined target frame.

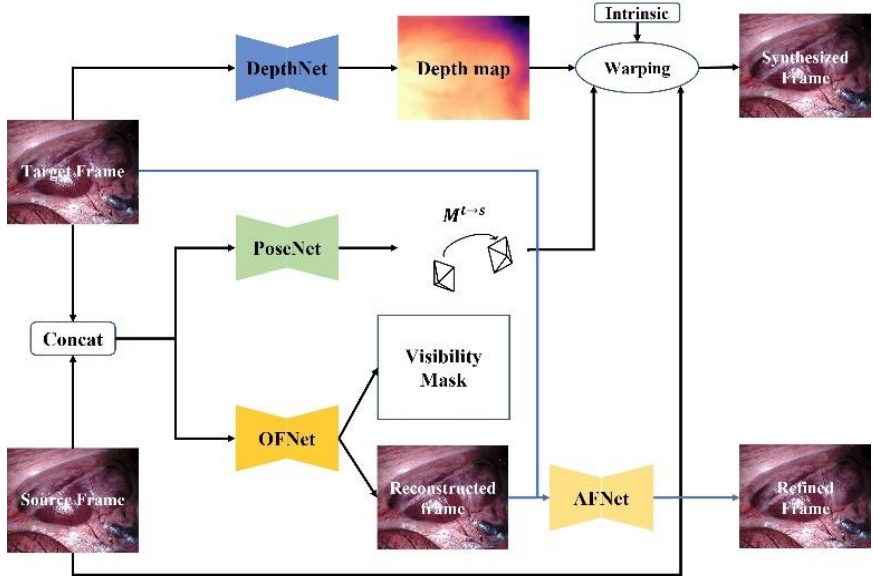

Fig. 2. Architecture of AF-SfMLearner.

The system is supervised by two losses as shown in equation 2.
$$\mathcal{L} = \mathcal{D}(\mathbf{p}) + \kappa \mathcal{R}(\mathbf{p}), \tag{2}$$
where $\mathcal{D}(\mathbf{p})$ and $\mathcal{R}(\mathbf{p})$ are data fidelity and a Tikhonov regularizer, respectively. And the $\kappa$ is loss weight to balance these two items.

$$\mathcal{D}(\mathbf{p}) = \sum_{\mathbf{p}} V(\mathbf{p}) * \Phi\left(I^{s \to t}(\mathbf{p}), I^t(\mathbf{p}) + C(\mathbf{p})\right) \tag{3}$$

where $V(\mathbf{p})$ is visibility mask generated by OFNet, $I^{s \to t}(\mathbf{p})$ is synthesized target frame. Here we choose two nearby frames of $I^t(\mathbf{p})$ as $I^s(\mathbf{p})$, $I^s(\mathbf{p}) \in \{I^{t-1}(\mathbf{p}), I^{t+1}(\mathbf{p})\}$. And $C(\mathbf{p})$ is the brightness calibration image which used to generate refined target frames. Finally, $\Phi(I^{s \to t}, I^t) = \alpha \frac{1 - SSIM(I^{s \to t}, I^t)}{2} + (1 - \alpha)|I^{s \to t} - I^t|_1$.

The Tikhonov regularizer $R(p)$ contains three parts as shown in equation 4.
$$R(\mathbf{p}) = \lambda_1 \mathcal{L}_{rs} + \lambda_2 \mathcal{L}_{ax} + \lambda_3 \mathcal{L}_{es}, \tag{4}$$
where $\mathcal{L}_{rs}$, $\mathcal{L}_{ax}$ and $\mathcal{L}_{es}$ are shown in equation 5~7.
$$\mathcal{L}_{rs} = \sum_{\mathbf{p}} |\nabla C(\mathbf{p})| * e^{-\nabla |I^t(\mathbf{p}) - I^{s \to t}(\mathbf{p})|}. \tag{5}$$

$\mathcal{L}_{rs}$ is defined as residual-based smoothness loss to encourage the AFNet generate smooth appearance flow map.
$$\mathcal{L}_{ax} = \sum_{\mathbf{p}} V(\mathbf{p}) * \Phi\left(I^{s \to t}(\mathbf{p}), I^t(\mathbf{p}) + C(\mathbf{p})\right). \tag{6}$$

$\mathcal{L}_{ax}$ is designed to provide auxiliary supervision for AFNet. The difference between $\mathcal{L}_{ax}$ and $\mathcal{D}(\mathbf{p})$ is $I^{s \to t}(\mathbf{p})$ represents the reconstructed target frames by optical flow from OFNet in $\mathcal{L}_{ax}$.
$$\mathcal{L}_{es} = \sum_{\mathbf{p}} |\nabla D(\mathbf{p})| * e^{-\nabla |I^t(\mathbf{p})|}. \tag{7}$$

The edge-aware smoothness loss $\mathcal{L}_{es}$ is designed to let predicted depth maps more smoothly.

### 3.2. Camera Intrinsic CNN

The architecture of camera intrinsic CNN is shown in Fig. 3. We choose the pre-trained ResNet-18 [23] as encoder, followed by a pose decoder and a camera decoder to predict the ego-motion and intrinsic matrix, respectively. In the camera decoder route, the feature from the encoder passes through a convolution layer and average pooling layer, then to the focal length convolution and offset convolution layer to generate the focal length and principal point offset of camera intrinsic parameters.

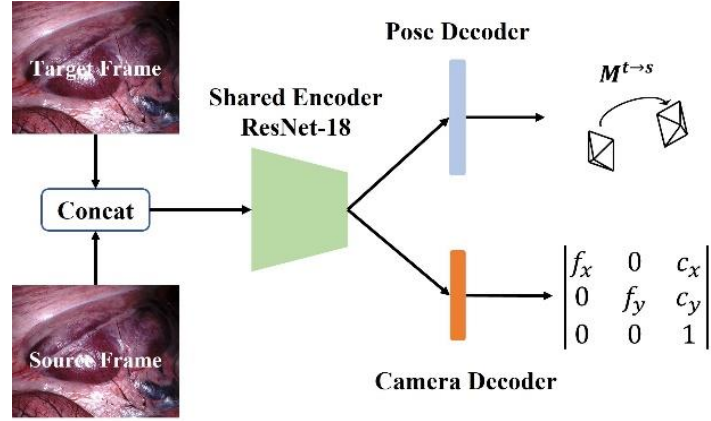

Fig. 3. Architecture camera intrinsic prediction CNN.

### 3.3. Cost Volume

Our cost volume module was inspired by ManyDepth [24], which aims to measure the geometric compatibility at different depth values between the pixels from target and source frames from the input videos. Our cost volume module architecture is shown in Fig. 4. It shares the same PoseNet and CameraNet with AF-SfMlearner to generate ego-motion and camera intrinsic parameters. Then we also use ResNet-18 as the encoder to extract the feature maps of target and source frames. In the cost volume computing, we only use past frames as nearby view of target frames, $I^s(\mathbf{p}) \in \{ I^{t-1}(\mathbf{p}), \ldots, I^{t-N}(\mathbf{p})\}$. Then we define a set of ordered depth planes $\mathcal{P}$, each perpendicular to the optical axis at $I^t$. The depth planes are linearly distributed between $d_{min}$ and $d_{max}$. After warping source feature maps with predicted ego-motion, intrinsic parameters, and defined depth plane to target domain, we calculate the $l_1$ distance between target feature map $f^t$ and synthesized target feature map $f^{s \to t}$ to build the cost volume. Then we use DepthNet to predict target depth map from the cost volume, which permits the network to leverage inputs from multiple viewpoints.

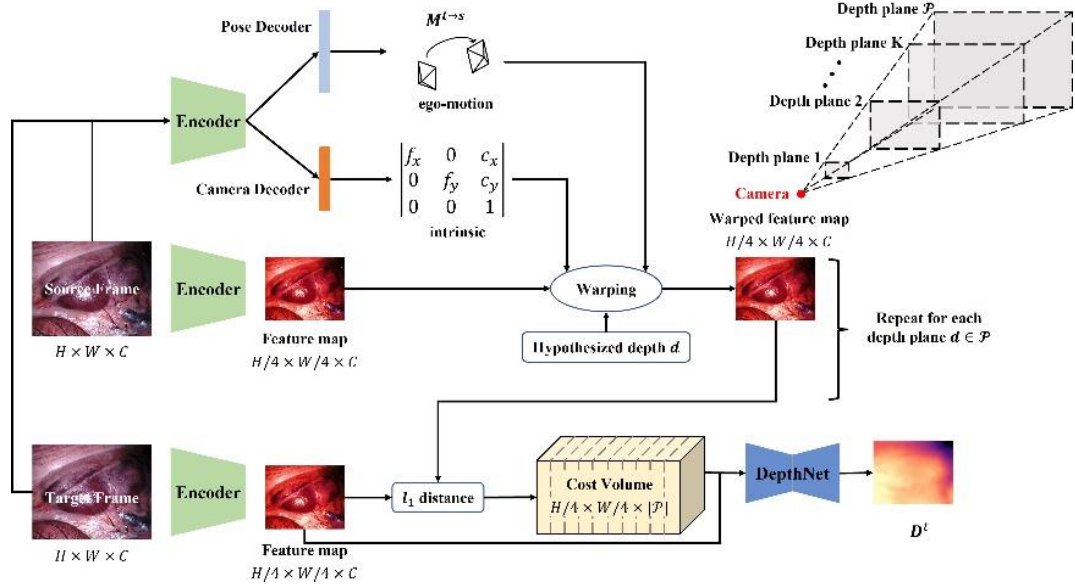

Fig. 4. Architecture of cost volume module.

### 3.4. WS-SfMLearner

In previous sections we introduce the architecture of each component. In this section, we will outline the overall structure of our WS-SfMLearner system as shown in Fig. 5. A pair of target and source frames are sent to the camera intrinsic module to obtain the predicted camera intrinsic matrix. And then these image pairs are fed into baseline and cost volume modules to generate depth maps $D_{sfm}^t$ and $D_c^t$, respectively. The whole system is supervised by the consistency loss between $D_{sfm}^t$ and $D_c^t$.

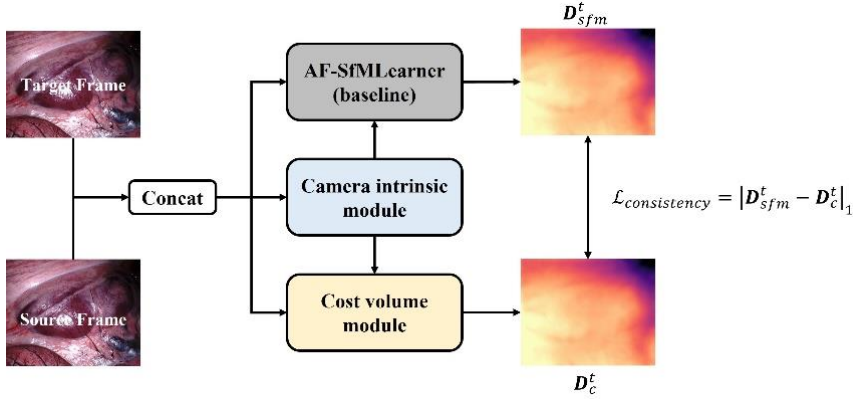

Fig.5. Architecture of WS-SfMLearner.

## 4. EXPERIMENTS

### 4.1. Implementation Details

The proposed model is developed in PyTorch [25]. All encoders in AF-SfMLearner, camera intrinsic module and cost volume module use ResNet-18 pretrained on ImageNet [26], and the architecture of DepthNet and PoseNet are the same as the classic SfMLearner. For baseline – AF-SfMLearner, we choose α κ, $\lambda_1$, $\lambda_2$ and $\lambda_3$ equal to 0.85, 1, 0.01, 0.001 and 0.0001, respectively. For the cost volume module, we set $d_{min} = 0.1$ and $d_{max} = 10$. We choose Adam optimizer with $\beta_1 = 0.9$ and $\beta_2 = 0.99$, batch size is 12 and initial learning rate is $1e-4$. And learning rate is multiplied by 0.1 after each 10 epochs. The total epochs we set to 20. We set the resolution of input image to 320×256. And the whole system is training on a single RTX A5000 GPU.

We test our novel method and compared with other state-of-the-arts such as SfMLeanrer [7], Monodepth2 [8], HR-Depth [27], Li et al [28], Li et al [29] (MICCAI'22) and AF-SfMLearner [20] on SCARED dataset [30]. The dataset contains 9 various scenes, and we divided it at 8:1:1 ratio for training, validation and testing based on video sequence.

### 4.2. Measurement Metrics

In the depth evaluation phase, we use the DepthNet from AF-SfMLearner to predict the final depth maps. Then we scaled the depth maps with median scaling from SfMLearner [7] which can be expressed as

$$D_{scaled} = \left( D_{pred} * \left( \frac{median(D_{gt})}{median(D_{pred})} \right) \right). \tag{8}$$

The scaled depth maps are capped at 200 mm on the SCARED dataset. The measurement metrics we used to evaluate the performance are listed below:

$$Abs\ Rel = \frac{1}{|D|} \sum_{d \in D} \frac{|d^* - d|}{d^*}, \tag{9}$$

$$Sq\ Rel = \frac{1}{|D|} \sum_{d \in D} \frac{|d^* - d|^2}{d^*}, \tag{10}$$

$$RMSE = \sqrt{\frac{1}{|D|} \sum_{d \in D} |d^* - d|^2}, \tag{11}$$

$$RMSE \log = \sqrt{\frac{1}{|D|} \sum_{d \in D} |\log d^* - \log d|^2}, \tag{12}$$

$$\delta = \frac{1}{|D|} \left| \left\{ d \in D \ \middle| \ \max\left(\frac{d^*}{d}, \frac{d}{d^*}\right) < 1.25 \right\} \right| \times 100\%. \tag{13}$$

$d$ and $d^*$ denote the predicted depth value and the corresponding ground truth, $D$ represents a set of predicted depth values.

## 4.3. Results

Table 1 shows the quantitative results for depth estimation. The depth estimation results show even with unknown camera intrinsic parameters, our WS-SfMLearner could predict high quality depth maps competitive with other state-of-the-art methods. In Table 2, we show the ablation studies for camera intrinsic and cost volume modules and report the ego-motion, camera intrinsic parameters and depth estimation results. It is easy to observe that if we directly insert the intrinsic prediction module into baseline without the cost volume, the performance declines on both depth and ego-motion estimation. After we use cost volume module to provide auxiliary supervision, the rotation and trajectory errors decrease to less than baseline with given intrinsic parameters. In addition, we also visualize the prediction trajectory in Fig. 6, and it is clear to show that our proposed method helps the system predict more accurate camera motion trajectories.

Table 1. Comparison of proposed model with existing models for depth estimation evaluation. "↑" and "↓" denotes higher and lower is better, respectively.

|  | Abs Rel↓ | Sq Rel↓ | RMSE↓ | RMSE log↓ | δ<1.25↑ | δ<$1.25^2$↑ | δ<$1.25^3$↑ |
|---|---|---|---|---|---|---|---|
| SfMLearner | 0.100 | 2.539 | 11.916 | 0.138 | 0.910 | 0.977 | 0.993 |
| Monodepth2 | 0.075 | 0.827 | 7.538 | 0.100 | 0.943 | 0.995 | 0.999 |
| HR-Depth | 0.076 | 0.864 | 7.718 | 0.103 | 0.941 | 0.995 | 1.000 |
| Li et al. [13] | 0.066 | 0.715 | 6.684 | 0.089 | 0.952 | 0.995 | 1.000 |
| Li et al. [14] | 0.062 | 0.654 | 6.649 | **0.086** | 0.956 | **0.997** | **1.000** |
| AF-SfMLearner | 0.065 | 0.575 | 5.763 | 0.093 | 0.957 | 0.995 | 0.999 |
| Ours | **0.062** | **0.565** | **5.754** | 0.089 | **0.968** | 0.995 | 0.999 |

Table 2. Ablation studies for proposed module for ego-motion, camera intrinsic parameters and depth evaluation. For ego motion and camera intrinsic parameters evaluation, we calculate the $l_2$ norm between prediction and ground truth, and results are shown in **mean ± std** form. For depth estimation, we report scores of five measurement metrics. (Note: Rotation, trajectory and camera intrinsic are normalized to [0,1] to calculate the accuracy, and the camera intrinsic ground truth is $f_x = 0.82, f_y = 1.02, c_x = c_y = 0.5$.)

| | Method | Baseline | Baseline + Camera | Baseline + Camera + Cost Volume |
|---|---|---|---|---|
| Ego motion | Rotation error | 0.0433±0.0232 | 0.0490±0.0234 | **0.0400±0.0221** |
| Ego motion | Trajectory error | 0.0871±0.0568 | 0.0996±0.0608 | **0.0737±0.0517** |
| Intrinsic | $f_x$ | - | 0.86±0.005 | **0.81±0.003** |
| Intrinsic | $f_y$ | - | 1.07±0.007 | **1.01±0.004** |
| Intrinsic | $c_x$ | - | 0.503±0.005 | **0.501±0.003** |
| Intrinsic | $c_y$ | - | 0.505±0.004 | **0.498±0.003** |
| Depth | Abs Rel↓ | 0.065 | 0.068 | **0.062** |
| Depth | Sq Rel↓ | 0.575 | 0.616 | **0.565** |
| Depth | RMSE↓ | 5.763 | 5.898 | **5.754** |
| Depth | RMSE log↓ | 0.093 | 0.095 | **0.089** |
| Depth | δ<1.25↑ | 0.957 | 0.952 | **0.968** |

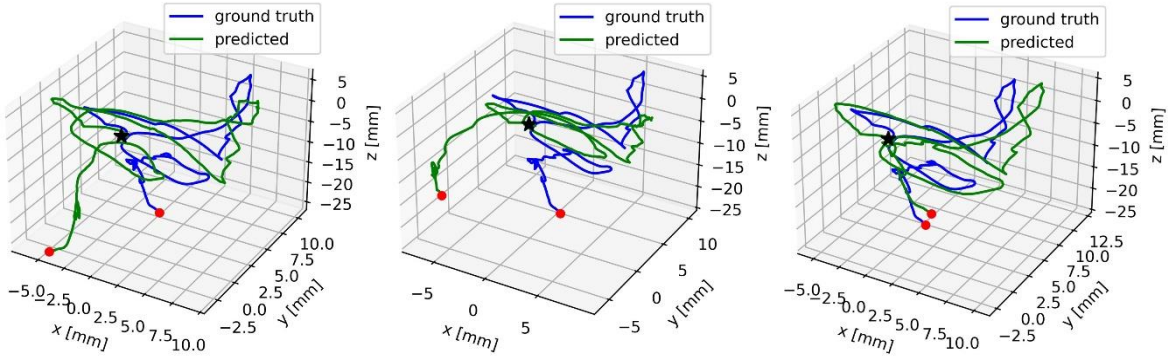

Fig 6. Visualization of trajectory. From left to right are trajectories predicted by baseline with given intrinsic, baseline with predicted intrinsic and our final version WS-SfMLearner.

## 5. CONCLUSIONS

We introduce the camera intrinsic prediction module in surgical scene depth estimation task to solve the problem of learning from unknown surgical videos. Next, we proposed a cost volume supervision approach. From the experimental results of depth, ego-motion and intrinsic prediction, the proposed cost volume supervision manner improves the robustness and accuracy of the SfMLearner-based system when learning from unknown surgical videos. The cost volume method can still be improved by learning depth plane distribution rather than using linear distribution directly such as using sinusoidal activation function [33], which will be considered in our future work. Additionally, addressing the challenge from moving objects, such as surgical tools within the scene, numerous segmentation methods [34-38] have been designed to segment and recognize medical objects. Incorporating these segmentation techniques into WS-SfMLearner to alleviate dynamic uncertainties and refine depth prediction precision for the surgical background would an interesting direction.